\documentclass{article}
\usepackage{iclr2017_conference,times}
\usepackage{times}
\usepackage{graphicx}
\usepackage{color}
\usepackage{amsmath,amssymb}
\usepackage[]{algorithm2e}

\newcommand{\xpt}{\edef\f@size{\@xpt}\rm}
\def\@xpt{10}

\newcommand{\cR}{{\cal R}}
\newcommand{\cA}{{\cal A}}
\newcommand{\cD}{{\cal D}}
\newcommand{\cO}{{\cal O}}
\newcommand{\cB}{{\cal B}}

\def\be {\begin{equation}}
\def\ee {\end{equation}}
\def\beas {\begin{eqnarray*}}
\def\eeas {\end{eqnarray*}}
\def\bea {\begin{eqnarray}}
\def\eea {\end{eqnarray}}
\def\bes {\begin{equation*}}
\def\ees {\end{equation*}}
\def\ba {\begin{align}}
\def\ea {\end{align}}
\def\barr {\begin{array}}
\def\earr {\end{array}}

\newtheorem{lemma-ap}{Lemma}

\newtheorem{claim-ap}{Claim}

\makeatletter
\usepackage{xspace}
\def\@onedot{\ifx\@let@token.\else.\null\fi\xspace}
\DeclareRobustCommand\onedot{\futurelet\@let@token\@onedot}

\newcommand{\figref}[1]{Fig\onedot~\ref{#1}}
\newcommand{\equref}[1]{Eq\onedot~\eqref{#1}}
\newcommand{\secref}[1]{Sec\onedot~\ref{#1}}
\newcommand{\tabref}[1]{Table~\ref{#1}}

\newcommand{\algref}[1]{Algorithm~\ref{#1}}

\def\eg{\emph{e.g}\onedot} 
\def\ie{\emph{i.e}\onedot} 
 
\def\etc{\emph{etc}\onedot} 
\def\wrt{w.r.t\onedot}

\makeatother

\begin{document}

\title{Learning to Play in a Day: Faster Deep Reinforcement Learning by Optimality Tightening}

\author{Frank S. He\\
Department of Computer Science\\
University of Illinois at Urbana-Champaign\\
Zhejiang University\\
\texttt{frankheshibi@gmail.com}
\And
Yang Liu\\
Department of Computer Science\\
University of Illinois at Urbana-Champaign\\
\texttt{liu301@illinois.edu}
\And
Alexander G. Schwing\\
Department of Electrical and Computer Engineering\\
University of Illinois at Urbana-Champaign\\
\texttt{aschwing@illinois.edu}
\And
Jian Peng\\
Department of Computer Science\\
University of Illinois at Urbana-Champaign\\
\texttt{jianpeng@illinois.edu}
}


%

\maketitle

\begin{abstract}
We propose a novel training algorithm for reinforcement learning which combines the strength of deep Q-learning with a constrained optimization approach to tighten optimality and encourage faster reward propagation.
Our novel technique makes deep reinforcement learning more practical by drastically reducing the training time.  We evaluate the performance of our approach on the 49 games of the challenging Arcade  Learning Environment, and report significant improvements in both training time and accuracy.
\end{abstract}
\section{Introduction}

The recent advances of supervised deep learning techniques~\citep{LeCunNature2015} in computer vision, speech recognition and natural language processing have tremendously improved the performance on challenging tasks, including image processing~\citep{KrizhevskyNIPS2012}, speech-based translation~\citep{SutskeverNIPS2014} and language modeling~\citep{HintonSPM2012}. The core idea of deep learning is to use artificial neural networks to model complex hierarchical or compositional data abstractions and representations from raw input data~\citep{BengioPAMI2013}.  However, we are still far from building intelligent solutions for many real-world challenges, such as autonomous driving, human-computer interaction and automated decision making, in which software agents need to consider interactions with a dynamic environment and take actions towards goals. Reinforcement learning~\citep{BertsekasNDP1996,PowellADP2011,SuttonMIT1998,KaelblingJMLR1996} studies these problems and algorithms which learn policies to make decisions so as to maximize a reward signal from the environment. One of the promising algorithms is Q-learning~\citep{WatkinsPHD1989,WatkinsML1992}.
Deep reinforcement learning with neural function approximation ~\citep{TsitsiklisTAC1997,RiedmillerECML2005,MnihNIPSWS2013,MnihNature2015}, possibly a first attempt to combine deep learning and reinforcement learning, has been proved to be effective on a few problems which classical AI approaches were unable to solve. Notable examples of deep reinforcement learning include human-level game playing~\citep{MnihNature2015} and AlphaGo~\citep{SilverNature2016}.

Despite these successes, its high demand of computational resources makes deep reinforcement learning not yet applicable to many real-world problems. For example, even for an Atari game, the deep Q-learning algorithm (also called deep Q-networks, abbreviated as DQN) needs to play up to hundreds of millions of game frames to achieve a reasonable performance~\citep{HasseltARXIV2015}. AlphaGo trained its model using a database of game records of advanced players and, in addition, about 30 million self-played game moves~\citep{SilverNature2016}. The sheer amount of required computational resources of current deep reinforcement learning algorithms is a major bottleneck for its applicability to real-world tasks. Moreover, in many tasks, the reward signal is sparse and delayed, thus making the convergence of learning even slower.

Here we propose optimality tightening, a new technique to accelerate deep Q-learning by fast reward propagation. While current deep Q-learning algorithms rely on a set of experience replays, they only consider a single forward step for the Bellman optimality error minimization, which becomes highly inefficient when the reward signal is sparse and delayed. To better exploit long-term high-reward strategies from past experience, we design a new algorithm to capture rewards from both forward and backward steps of the replays via a constrained optimization approach. This encourages faster reward propagation which reduces the training time of deep Q-learning.

We evaluate our proposed approach using the Arcade learning environment~\citep{BellemareJAIR2013} and show that our new strategy  outperforms competing techniques in both accuracy and training time on  30 out of 49 games despite being trained with significantly fewer data frames.
\section{Related Work}
There have been a number of approaches improving the stability, convergence and runtime of deep reinforcement learning since  deep Q-learning, also known as deep Q-network (DQN), was first proposed~\citep{MnihNIPSWS2013,MnihNature2015}. DQN combined techniques such as deep learning, reinforcement learning and experience replays~\citep{LinML1992,WawrzynskiNN2009}.

Nonetheless, the original DQN algorithm required millions of training steps to achieve human-level performance on Atari games. To improve the stability, recently, double Q-learning was combined with deep neural networks, with the goal to alleviate the overestimation issue observed in Q-learning~\citep{ThrunConnectionist1993,HasseltNIPS2010,HasseltARXIV2015}. The key idea is to use two Q-networks for the action selection and Q-function value calculation, respectively. The greedy action of the target is first chosen using the current Q-network parameters, then the target value is computed using a set of parameters from a previous iteration. Another notable advance is ``prioritized experience replay''~\citep{SchaulICLR2016} or ``prioritized sweeping'' for deep Q-learning. The idea is to increase the replay probability of experience tuples that have a high expected learning progress measured by temporal difference errors. 

In addition to the aforementioned variants of Q-learning, other network architectures have  been proposed. The dueling network architecture applies an extra network structure to learn the importance of states and uses advantage functions~\citep{WangARXIV2015}. A distributed version of the deep actor-critic algorithm without experience replay was introduced very recently~\citep{MnihARXIV2016}. It deploys multiple threads learning directly from current transitions. The approach is applicable to both value-based and policy-based methods, off-policy as well as on-policy methods, and in discrete as well as in continuous domains. The model-free episodic control approach evaluates state-action pairs based on episodic memory using k-nearest neighbors with hashing functions~\citep{BlundellARXIV2016}. Bootstrapped deep Q-learning carries out temporally-extended (or deep) exploration, thus leading to much faster learning~\citep{OsbandARXIV2016}.

Our fast reward propagation differs from all of the aforementioned approaches. The key idea of our method is to propagate delayed and sparse rewards during Q-network training, and thus greatly improve the efficiency and  performance. We formulate this propagation step via a constrained program. Note that our program is also different from earlier work on off-policy $Q^\ast(\lambda)$ algorithms with eligibility traces ~\citep{MunosNIPS2016,WatkinsPHD1989}, which have been recently shown to perform poorly when used for training deep Q-networks on Atari games.

\section{Background}

Reinforcement learning considers agents which are able to take a sequence of actions in an environment. By taking actions and experiencing at most one scalar reward per action, their task is to learn a policy which allows them to act such that a high cumulative reward is obtained over time.

More precisely, consider an agent operating over time $t\in\{1, \ldots, T\}$. At time $t$ the agent is in an environment state $s_t$ and reacts upon it by choosing action $a_t\in\cA$. The agent will then observe a new state $s_{t+1}$ and receive a numerical reward $r_t \in \cR$. Throughout, we assume the set of possible actions, \ie, the set $\cA$, to be discrete.

A well established technique to address the aforementioned reinforcement learning task is Q-learning~\citep{WatkinsPHD1989,WatkinsML1992}. Generally, Q-learning algorithms maintain an action-value function, often also referred to as Q-function, $Q(s,a)$. Given a state $s$, the action-value function provides a `value' for each action $a\in\cA$ which estimates the expected future reward if action $a\in\cA$ is taken. The estimated future reward is computed based on the current state $s$ or a series of past states $s_t$ if available.

The core idea of Q-learning is the use of the Bellman equation as a characterization of the optimal future reward function $Q^\ast$ via a state-action-value function
\be
Q^\ast(s_t,a) = \mathbb{E}[r_t + \gamma\max_{a^\prime} Q^\ast(s_{t+1},a^\prime)].
\label{eq:Bellman}
\ee
Hereby the expectation is taken \wrt the distribution of state $s_{t+1}$ and reward $r_t$ obtained after taking action $a$, and $\gamma$ is a discount factor.  Intuitively, reward for taking action $a$ plus best future reward should equal the best total return from the current state.

The choice of Q-function is crucial for the success of Q-learning algorithms. While classical methods use linear Q-functions based on a set of hand-crafted features of the state, more recent approaches use nonlinear deep neural networks to automatically mine intermediate features from the state~\citep{RiedmillerECML2005,LangeConfNN2010,MnihNIPSWS2013,MnihNature2015}. This change has been shown to be very effective for many applications of reinforcement learning. However, automatic mining of intermediate representations comes at a price: larger quantities of data and more computational resources are required. Even though it is sometimes straightforward to extract large amounts of data, \eg, when training on  video games, for successful optimization, it is crucial that the algorithms operate on un-correlated samples from a dataset $\cD$ for stability. A technique called ``experience replay''~\citep{LinML1992,WawrzynskiNN2009} encourages this property and quickly emerged as a standard step in the well-known deep Q-learning framework~\citep{MnihNIPSWS2013,MnihNature2015}. Experience replays are stored as a dataset $\cD = \{(s_j, a_j,  r_j, s_{j+1})\}$ which contains state-action-reward-future state-tuples $(s_j, a_j, r_j, s_{j+1})$, including past observations from previous plays. 

The characterization of optimality given in \equref{eq:Bellman} combined with an ``experience replay'' dataset $\cD$ results in the following iterative algorithmic procedure~\citep{MnihNIPSWS2013,MnihNature2015}: 
 start an episode in the initial state $s_0$; sample a mini-batch of tuples $\cB=\{(s_j, a_j, r_j, s_{j+1})\}\subseteq\cD$; compute and fix the targets $y_j = r_j + \gamma\max_a Q_{\theta^-}(s_{j+1}, a)$ for each tuple using a recent estimate $Q_{\theta^-}$ (the maximization is only considered if $s_j$ is not a terminal state); update the Q-function by optimizing the following program \wrt the parameters $\theta$ typically via stochastic gradient descent: 
\be
\min_\theta \sum_{(s_j, a_j, r_j, s_{j+1})\in\cB}\left(Q_{\theta}(s_j,a_j) - y_j\right)^2.
\label{eq:QFuncOpt}
\ee
After having updated the parameters of the Q-function we perform an action simulation either choosing an action at random with a small probability $\epsilon$, or by following the strategy $\arg\max_{a} Q_{\theta}(s_t, a)$ which is currently estimated. This strategy is also called the $\epsilon$-greedy policy. We then obtain the actual reward $r_t$. Subsequently we augment the replay memory with the new tuple $(s_t, a_t,r_t, s_{t+1})$ and continue the simulation until this episode terminates or reaches an upper limit of steps, and we restart a new episode. When optimizing \wrt the parameter $\theta$, a recent Q-network is used to compute the target $y_j = r_j + \gamma\max_a Q_{\theta^-}(s_{j+1},a)$. This technique is referred to as `semi-gradient descent,' \ie, the dependence of the target on the parameter $\theta$ is ignored.

\section{Fast Reward Propagation via Optimality Tightening}
Investigating the cost function given in \equref{eq:QFuncOpt} more carefully, we observe that it operates on a set of short one-step sequences, each characterized by the tuple $(s_j, a_j, r_j, s_{j+1})$. Intuitively, each step encourages an update of the parameters $\theta$, such that the action-value function for the chosen action $a_j$, \ie, $Q_\theta(s_j, a_j)$, is closer to the obtained reward plus the best achievable future value, \ie, $y_j = r_j + \gamma\max_a Q(s_{j+1},a)$. As we expect from the Bellman optimality equation, it is instructive to interpret this algorithm as propagating reward information from time $j+1$ backwards to time $j$.


To understand the shortcomings of this procedure consider a situation where the agent only receives a sparse and delayed reward once reaching a target in a maze. Further let $|P|$ characterize the shortest path from the agents initial position to the target. For a long time, no real reward is available and the aforementioned algorithm propagates randomly initialized future rewards. Once the target is reached, real reward information is available. Due to the cost function and its property of propagating reward time-step by time-step, it is immediately apparent that it takes at least an additional $\cO(|P|)$ iterations until the observed reward impacts the initial state.

In the following we propose a technique which increases the speed of propagation and achieves improved convergence for deep Q-learning. We achieve this improvement by taking advantage of longer state-action-reward-sequences which are readily available in the ``experience replay memory.'' Not only do we propagate information from time instances in the future to our current state, but also will we pass information from states several steps in the past. Even though we expect to see substantial improvements on sequences where rewards are sparse or only available at terminal states, we also demonstrate significant speedups for situations where rewards are obtained frequently. This is intuitive as the Q-function represents an estimate for any reward encountered in the future. Faster propagation of future and past rewards to a particular state is therefore desirable.

Subsequently we discuss our technique for fast reward propagation, a new deep Q-learning algorithm that exploits longer state-transitions in  experience replays by tightening the optimization via constraints.
From the Bellman optimality equation we know that the following series of equalities hold for the optimal Q-function $Q^\ast$:
\begingroup\makeatletter\def\f@size{9}\check@mathfonts
\bes
Q^\ast(s_j,a_j) = r_j + \gamma\max_a Q^\ast(s_{j+1},a) = \ldots = 
r_j + \gamma\max_a \left[r_{j+1} + \gamma\max_{a^\prime}\left[r_{j+2} + \gamma\max_{\tilde{a}} Q^\ast(s_{j+3},\tilde{a})\right]\right].
\ees
\endgroup
Evaluating such a sequence exactly is not possible in a reinforcement learning setting since the enumeration of intermediate states $s_{j+i}$ requires exponential time complexity $\cO(|\cA|^i)$. 
It is however possible to take advantage of the episodes available in the replay memory $\cD$ by noting that
 the following
 sequence of inequalities holds for the optimal action-value function $Q^\ast$ (with the greedy policy), irrespective of whether the policy generating the sequence of actions $a_j$, $a_{j+1}$, \etc, which results in rewards $r_j$, $r_{j+1}$, \etc is optimal or not:
\bes
Q^\ast(s_j,a_j) = r_j + \gamma\max_a Q^\ast(s_{j+1},a)\geq \ldots \geq\sum_{i=0}^k \gamma^i r_{j+i} + \gamma^{k+1}\max_a Q^\ast(s_{j+k+1},a) = L_{j,k}^\ast.
\ees
Note the definition of the lower bounds $L_{j,k}^\ast$ for sample $j$ and time horizon $k$ in the aforementioned series of inequalities.



We can also use this series of inequalities to  define upper bounds. To see this note that 
$$
Q^\ast(s_{j-k-1},a_{j-k-1}) \geq \sum_{i=0}^k \gamma^i r_{j-k-1+i} + \gamma^{k+1} Q^\ast(s_{j},a_{j}),
$$
which follows from the definition of the lower bound by dropping the maximization over the actions, and a change of indices from $j\rightarrow j-k-1$. Reformulating the inequality yields an upper bound $U_{j,k}^\ast$ for sample $j$ and time horizon $k$ as follows:
$$
U_{j,k}^\ast = \gamma^{-k-1}Q^\ast(s_{j-k-1},a_{j-k-1}) - \sum_{i=0}^k \gamma^{i-k-1} r_{j-k-1+i} \geq Q^\ast(s_j,a_j).
$$

In contrast to classical techniques which optimize the Bellman criterion given in \equref{eq:QFuncOpt}, we propose to optimize the Bellman equation subject to constraints $Q_\theta(s_j,a_j) \geq L_j^{\max} = \max_{k\in\{1,\ldots,K\}}L_{j,k}$, which defines the largest lower bound, and $Q_\theta(s_j,a_j)\leq U_j^{\min} = \min_{k\in\{1, \ldots, K\}} U_{j,k}$, which specifies the smallest upper bound. Hereby, $L_{j,k}$ and $U_{j,k}$ are computed using the Q-function $Q_{\theta^-}$ with a recent estimated parameter $\theta^-$ rather than the unknown optimal Q-function $Q^\ast$, and the integer $K$ specifies the number of future and past time steps which are considered. Also note that the target used in the Bellman equation is obtained from $y_j = L_{j,0} = r_j + \gamma\max_a Q_{\theta^-}(s_{j+1},a)$. In this way, we ignore the dependence of the bounds and the target on the parameter $\theta$ to stabilize the training.
Taking all the aforementioned definitions into account, we propose  the following program for reinforcement learning tasks:
\be
\min_\theta \sum_{(s_j, a_j, s_{j+1}, r_j)\in\cB}\left(Q_{\theta}(s_j,a_j) - y_j\right)^2\quad\text{s.t.}\quad \left\{\begin{array}{cc}
Q_\theta(s_j,a_j) \geq L_j^{\max} & \forall~(s_j,a_j)\in\cB\\
Q_\theta(s_j,a_j) \leq U_j^{\min} & \forall~(s_j,a_j)\in\cB
\end{array}\right..
\label{eq:OurProgram}
\ee
This program differs from the classical approach given in \equref{eq:QFuncOpt} via the constraints, which is crucial. Intuitively, the constraints encourage faster reward propagation as we show next, and result in tremendously better results as we will demonstrate empirically in \secref{sec:exp}. 

\begin{figure}[t]
\fbox{
\begin{minipage}[c]{13.6cm}

\begin{algorithm}[H]
 \SetKwInOut{Init}{Initialize}
 \SetKwInOut{Output}{Output}
 \Output{Parameters $\theta$ of a Q-function}
 \Init{$\theta$ randomly, set $\theta^-=\theta$}
\For{$\operatorname{episode}\leftarrow 1$ \KwTo $M$}{
initialize $s_1$\;
\For{$t\leftarrow 1$ \KwTo $T$}{
Choose action $a_t$ according to $\epsilon$-greedy strategy\;
Observe reward $r_t$ and next state $s_{t+1}$\;
Store the tuple $(s_t,a_t,r_t,\cdot,s_{t+1})$ in replay memory $\cD$\;
Sample a minibatch of tuples $\cB=\{(s_j,a_j,r_j,R_j,s_{j+1}\})$ from replay memory $\cD$\;
Update $\theta$ with one gradient step of cost function given in  \equref{eq:OurFinalCost}\;
Reset $\theta^-=\theta$ every $C$ steps\;
}
\For{$t\leftarrow T$ \KwTo $1$}{
Compute $R_t = r_{t} + \gamma R_{t+1} $\;
Insert $R_t$ into the corresponding tuple in replay memory $\cD$\;
}

}

\caption{Our algorithm for fast reward propagation in reinforcement learning tasks.}
 \label{alg:OurAlgo}
\end{algorithm}
\vspace{-0.6cm}
\end{minipage}
}
\vspace{0.3cm}
\end{figure}

Before doing so we describe our optimization procedure for the constrained program in \equref{eq:OurProgram} more carefully. The cost function is generally non-convex in the parameters $\theta$, and so are the constraints. We therefore make use of a quadratic penalty method to reformulate the program into
\be
\min_\theta \hspace{-0.4cm}\sum_{(s_j, a_j, r_j, s_{j+1})\in\cB}\hspace{-0.4cm}\left[\left(Q_{\theta}(s_j,a_j) - y_j\right)^2 + \lambda(L_j^{\max}-Q_\theta(s_j,a_j))^2_+  + \lambda( Q_\theta(s_j,a_j)-U_j^{\min})_+^2\right],
\label{eq:OurFinalCost}
\ee
where $\lambda$ is a penalty coefficient and $(x)_+ = \max(0,x)$ is the rectifier function. Augmenting the cost function with $\lambda(L_j^{\max}-Q_\theta(s_j,a_j))^2_+$ and/or $\lambda( Q_\theta(s_j,a_j)-U_j^{\min})^2_+$ results in a penalty whenever any optimality bounding constraint gets violated. The quadratic penalty function is chosen for simplicity. The penalty coefficient $\lambda$ can be set as a large positive value or adjusted in an annealing scheme during training. In this work, we fix its value, due to time constraints. We optimize this cost function with stochastic (sub-)gradient descent using an experience replay memory from which we randomly draw samples, as well as their successors and predecessors. We emphasize that the derivatives correcting the prediction of $Q(s_j,a_j)$ not only depend on the Q-function from the immediately successive time step $Q(s_{j+1},a)$ stored in the experience replay memory, but also on more distant time instances if constraints are violated. Our proposed formulation and the resulting optimization technique hence encourage faster reward propagation, and the number of time steps depends on the constant $K$ and the quality of the current Q-function. We summarize the proposed method in \algref{alg:OurAlgo}.

The computational complexity of the proposed approach increases with the number of considered time steps $K$, since additional forward passes are required to compute the bounds $L_j^{\max}$ and $U_j^{\min}$. 
However, we can increase the memory size on the GPU to compute both the bounds and targets in a single forward pass  if $K$ is not too large.
If at all a problem, we can further alleviate this increase by randomly sampling a subset of the constraints rather than exhaustively using all of them. More informed strategies regarding the choice of constraints are possible as well since we may expect lower bounds in the more distant future to have a larger impact early in the training. In contrast once the algorithm is almost converged we may expect lower bounds close to the considered time-step to have bigger impact. 

To efficiently compute the discounted reward over multiple time steps we add a new element to the experience replay structure. Specifically, in addition to state, action, reward and next state for time-step $j$, we also store the real discounted return $R_j$ which is the discounted cumulative return achieved by the agent in its game episode. $R_j$ is computed via   
$
R_j = \sum_{\tau=j}^{T} \gamma^{\tau-j} r_{\tau}
$,
where $T$ is the end of the episode and $\gamma$ is the discount factor. $R_j$ is then inserted in the replay memory after the termination of the current episode or after reaching the limit of steps. All in all, the structure of our experience replay memory consists of tuples of the form $(s_j, a_j, r_j, R_j, s_{j+1})$. 

We leave the questions regarding a good choice of penalty function and a good choice of the penalty coefficients to future work. At the moment we use a  quadratic penalty function and a constant penalty coefficient $\lambda$ identical for both bounds. More complex penalty functions and sophisticated optimization approaches may yield even better results than the ones we report in the following.
\section{Experiments}
\label{sec:exp}
We evaluate the proposed algorithm on a set of 49 games from the Arcade Learning Environment~\citep{BellemareJAIR2013} as suggested by \citet{MnihNature2015}.  This environment is considered to be one of the most challenging reinforcement learning task because of its high dimensional output. Moreover, the intrinsic mechanism varies tremendously for each game, making it extremely demanding to find a single, general and robust algorithm and a corresponding single hyperparameter setting which works well across all 49 games.

Following existing work~\citep{MnihNature2015}, our agent predicts an action based on only raw image pixels and reward information received from the environment. 
A deep neural network is used as the function approximator for the Q-function. The game image is resized to an $84 \times 84$ grayscale image $s_t$. The first layer is a convolutional layer with 32 filters of size $8 \times 8$ and a stride of 4; the second layer is a convolutional layer with 64 filters of size $4\times 4$ and stride of 2; the third layer is a convolutional layer with 64 filters of size $3\times 3$ and a stride of 1; the next fully connected layer transforms the input to  512 units which are then transformed by another fully connected layer to an output size equal to the number of actions in each game. The rectified linear unit (ReLU) is used as the activation function for each layer. We used the hyperparameters provided by \citet{MnihNature2015} for annealing $\epsilon$-greedy exploration and also applied RMSProp for gradient descent. As in previous work we combine four frames into a single step for processing. We chose the hyperparamenter $K = 4$, for GPU memory efficiency when dealing with mini-batches. In addition, we also incorporate the discounted return $R_j$ in the lower bound calculation to further stabilize the training. We use the penalty coefficient $\lambda=4$ which was obtained by coarsely tuning performance on the games `Alien,' `Amidar,' `Assault,' and `Asterix.'  Gradients are also rescaled so that their magnitudes are comparable with or without penalty. All experiments are performed on an NVIDIA GTX Titan-X 12GB graphics card.

 \vspace{-0.5cm}
\begin{figure}[t!]
    \centering
    \includegraphics[width=\linewidth]{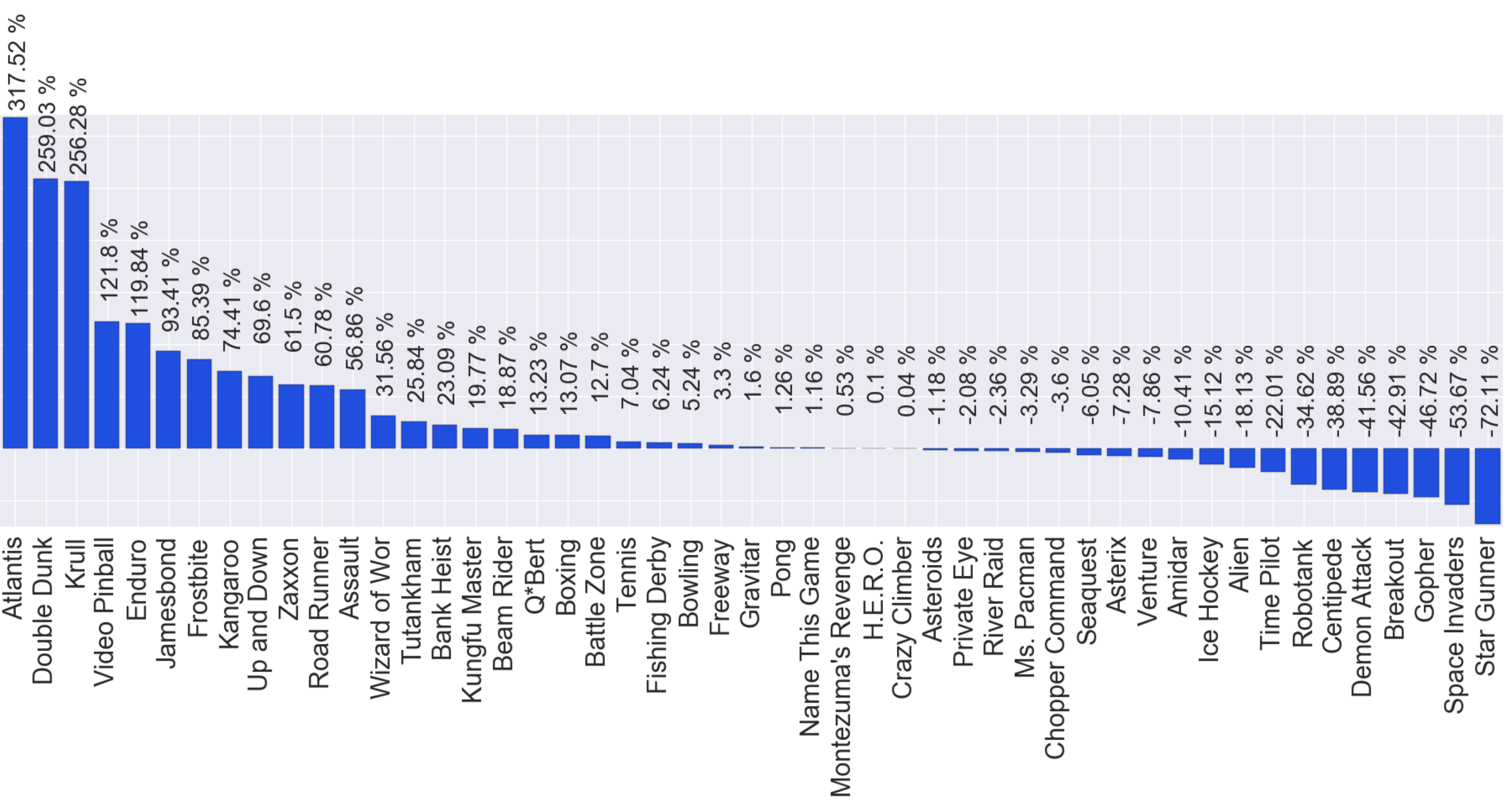}
    \caption{Improvements of our method trained on 10M frames compared to results of 200M frame DQN training presented by~\citet{MnihNature2015}, using the metric given in \equref{eq:evaluation1}.}
    \label{fig:Ours_with_nature200M_vertical}
    \vspace{-0.5cm}
\end{figure}

\subsection{Evaluation}
In previous work~\citep{MnihNature2015,HasseltARXIV2015,SchaulICLR2016,WangARXIV2015}, the Q-function is trained on each game using 200 million (200M) frames or 50M training steps. 
We compare to those baseline results obtained after 200M frames using our proposed algorithm which ran for only 10M frames or 2.5M steps, \ie, 20 times fewer data, due to time constraints. Instead of training more than 10 days we manage to finish training in less than one day. 
Furthermore, for a fair comparison, we replicate the DQN results and compare the performance of the proposed algorithm after 10M frames to those obtained when training DQN on only 10M frames.

\begin{figure}[t!]
    \centering
\vspace{-0.6cm}    \includegraphics[width=\linewidth]{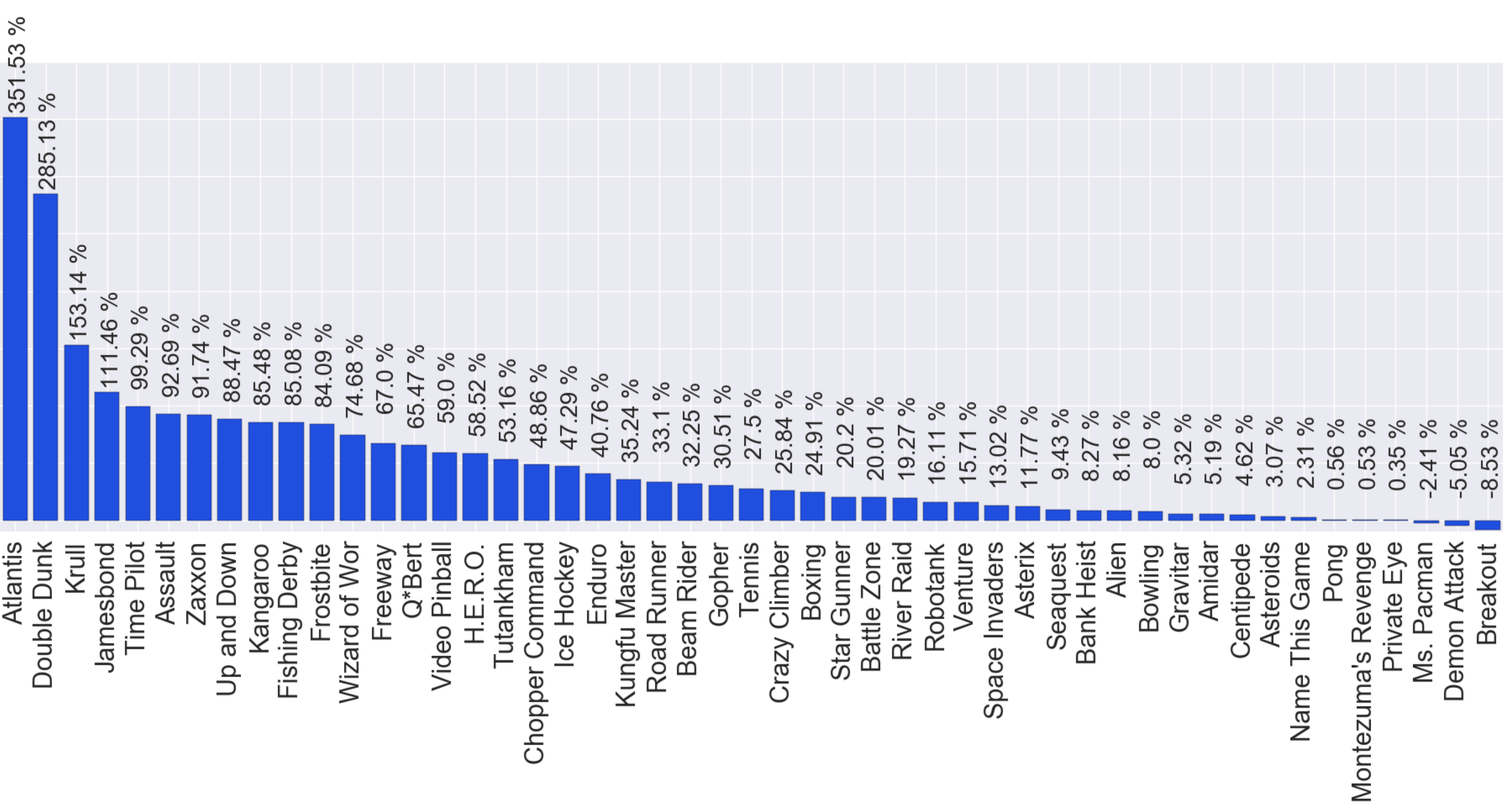}
    \caption{Improvements of our method trained on 10M frames compared to results of 10M frame DQN training, using the metric given in \equref{eq:evaluation1}.}    \label{fig:Ours_with_ourBaseline2_vertical2}
\vspace{-0.3cm}
\end{figure}

We strictly follow the evaluation procedure in~\citep{MnihNature2015} which is often referred to as `30 no-op evaluation.' During both training and testing, at the start of the episode, the agent always performs a random number of at most 30 no-op actions. During evaluation, our agent plays each game 30 times for up to 5 minutes, and the obtained score is averaged over these 30 runs. An $\epsilon$-greedy policy with $\epsilon =0.05$ is used. Specifically, for each run, the game episode starts with at most 30 no-op steps, and ends with `death' or after a maximum of 5 minute game-play, which corresponds to 18000 frames.

Our training consists of $M = 40$ epochs, each containing 250000 frames, thus 10M frames in total. For each game, we evaluate our agent at the end of every epoch, and, following common practice~\citep{HasseltARXIV2015,MnihNature2015}, we select the best agent's evaluation as the result of the game. So almost all hyperparameters are selected identical to \citet{MnihNature2015} and \citet{NairARIXV2015}.

To compare the performance of our algorithm to the DQN  baseline, we follow the approach of \citet{WangARXIV2015} and measure the improvement in percent using
\be\label{eq:evaluation1}
    \frac{\text{Score}_\text{Agent} - \text{Score}_\text{Baseline}}{\max \{\text{Score}_\text{Human}, \text{Score}_\text{Baseline}\} - \text{Score}_\text{Random}}.
\ee
We select this approach because the denominator choice of either human or baseline score prevents insignificant changes or negative scores from being interpreted as large improvements.

\figref{fig:Ours_with_nature200M_vertical} shows the improvement of our algorithm over the DQN baseline proposed by \citet{MnihNature2015} and trained for 200M frames, \ie, 50M steps. Even though our agent is only trained for 10M frames, we observe that our technique  outperforms the baseline significantly. In 30 out of 49 games, our algorithm exceeds the baseline using only $5\%$ of the baseline's training frames, sometimes drastically, \eg, in games such as `Atlantis,' `Double Dunk,' and `Krull.'    The remaining 19 games, often require a long training time. Nonetheless, our algorithm still reaches a satisfactory level of performance. 

\begin{table} 
\vspace{-0.3cm}
\centering
    \begin{tabular}{|c||c|c c|}
        \hline
        & Training Time & Mean & Median\\\hline\hline
        Ours (10M) & {\bf less than 1 day (1 GPU)} & {\bf 345.70\%} & {\bf 105.74\%}\\\hline
        DQN (200M) & more than 10 days (1 GPU) & 241.06\% & 93.52\%\\\hline\hline
        D-DQN (200M) & more than 10 days (1 GPU) & 330.3\% & {\bf 114.7\%}\\\hline
    \end{tabular}
    \caption{Mean and median human-normalized scores. DQN baseline and D-DQN results are from \cite{MnihNature2015,HasseltARXIV2015} and trained with 200M frames while our method is trained with 10M frames. Note that our approach can be combined with the D-DQN method.}
    \label{statistics_table1}
    \vspace{-0.3cm}
\end{table}

In order to further illustrate the effectiveness of our method, we compare our  results  with our implementation of DQN trained on 10M frames. The results are illustrated in \figref{fig:Ours_with_ourBaseline2_vertical2}.  We observe a better performance on 46 out of 49 games, demonstrating in a fair way the potential of our technique.



As suggested by~\cite{HasseltARXIV2015}, we use the following score
\begin{equation}\label{eq:evaluation3}
    \text{Score}_\text{Normalized} = \frac{\text{Score}_\text{Agent} - \text{Score}_\text{Random}}{\left | \text{Score}_\text{Human} - \text{Score}_\text{Random}\right |}
\end{equation}
to summarize the performance of our algorithm in a single number. We normalize the scores of our algorithm, the baseline reported by~\citet{MnihNature2015}, and double DQN (D-DQN)~\citep{HasseltARXIV2015}, and report the training time, mean and median in \tabref{statistics_table1}. We observe our technique with 10M frames to achieve comparable scores to the D-DQN method trained on 200M frames~\citep{HasseltARXIV2015}, while it outperforms the DQN method~\citep{MnihNature2015} by a large margin. We believe that our method can be readily combined with other techniques developed for DQN, such as D-DQN~\citep{HasseltARXIV2015}, prioritized experience replay~\citep{SchaulICLR2016}, dueling networks~\citep{WangARXIV2015}, and asynchronous methods~\citep{MnihARXIV2016} to further improve the accuracy and training speed.

\begin{figure}
    \centering
    \begin{tabular}{ccc}
    	\includegraphics[width=0.3\textwidth]{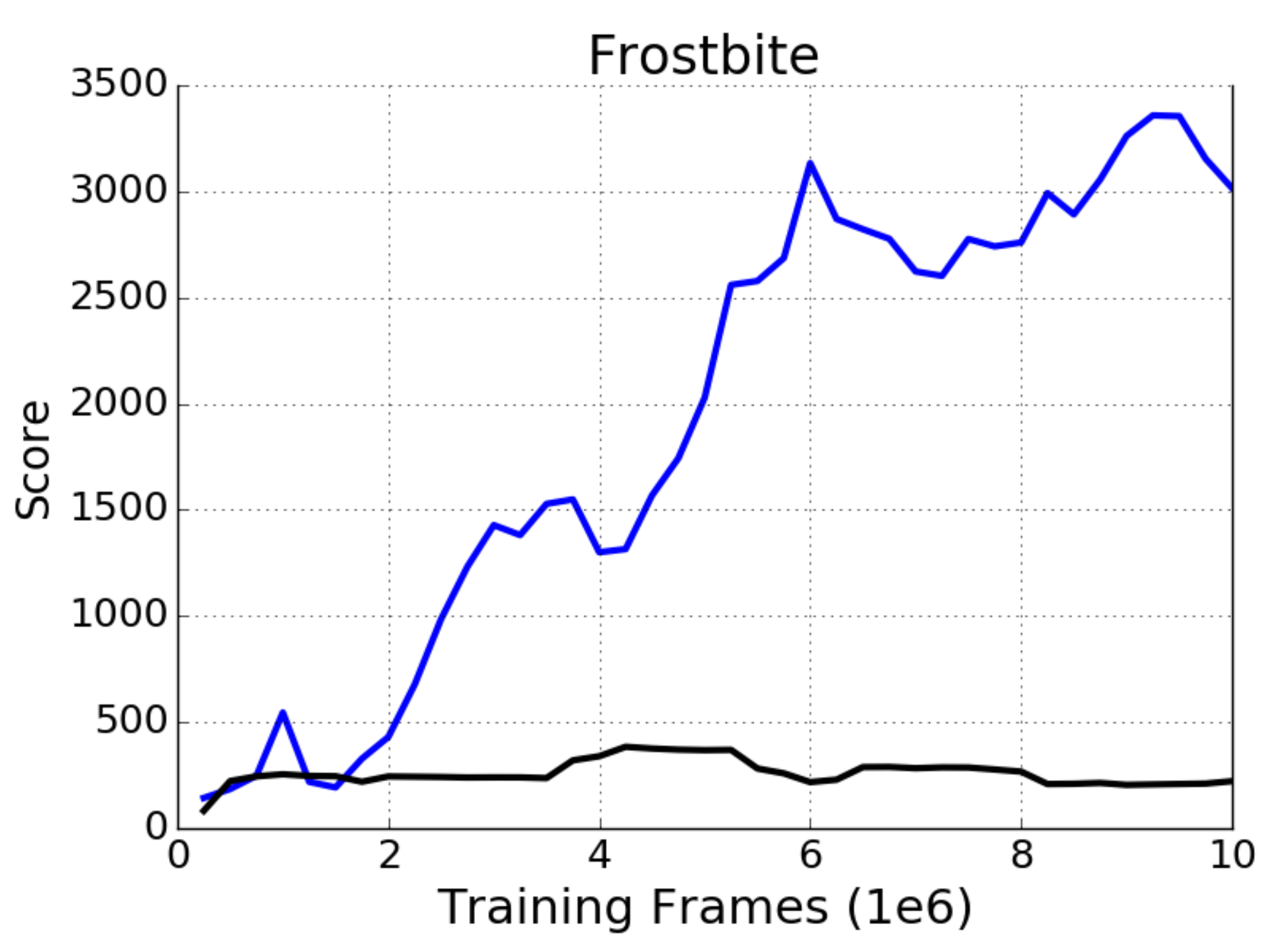}&
	\includegraphics[width=0.3\textwidth]{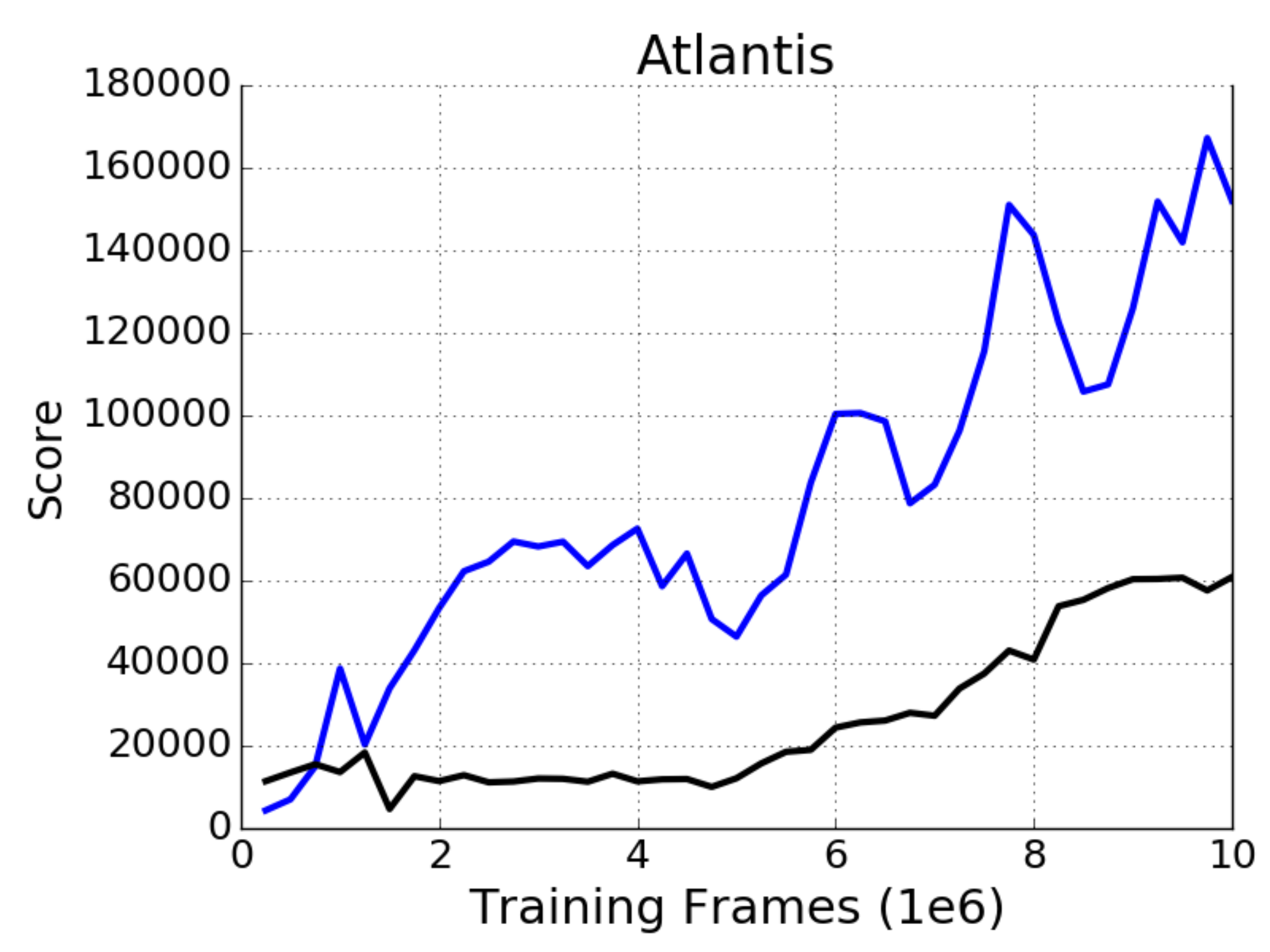}&
	\includegraphics[width=0.3\textwidth]{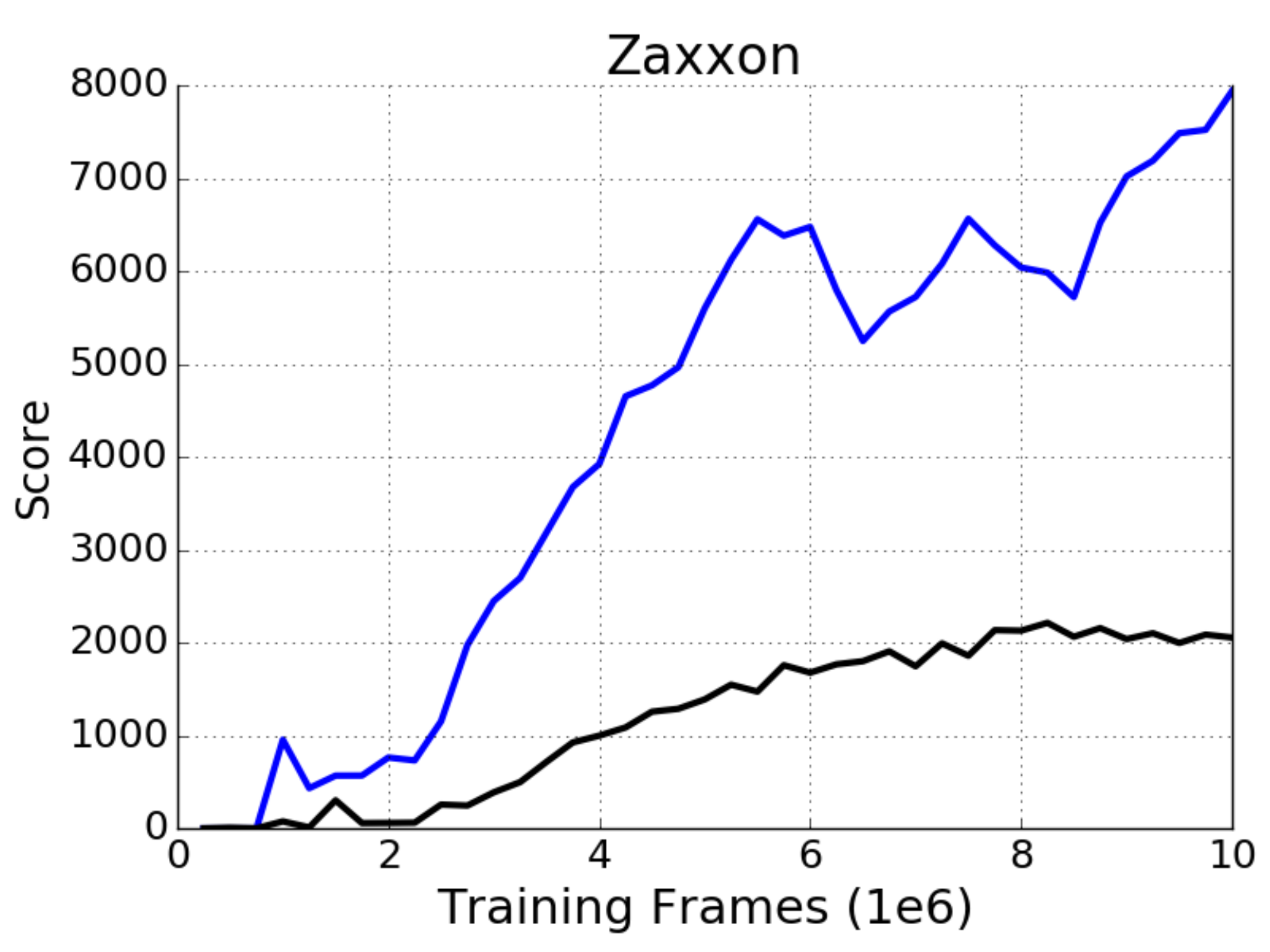}\\
	\includegraphics[width=0.3\textwidth]{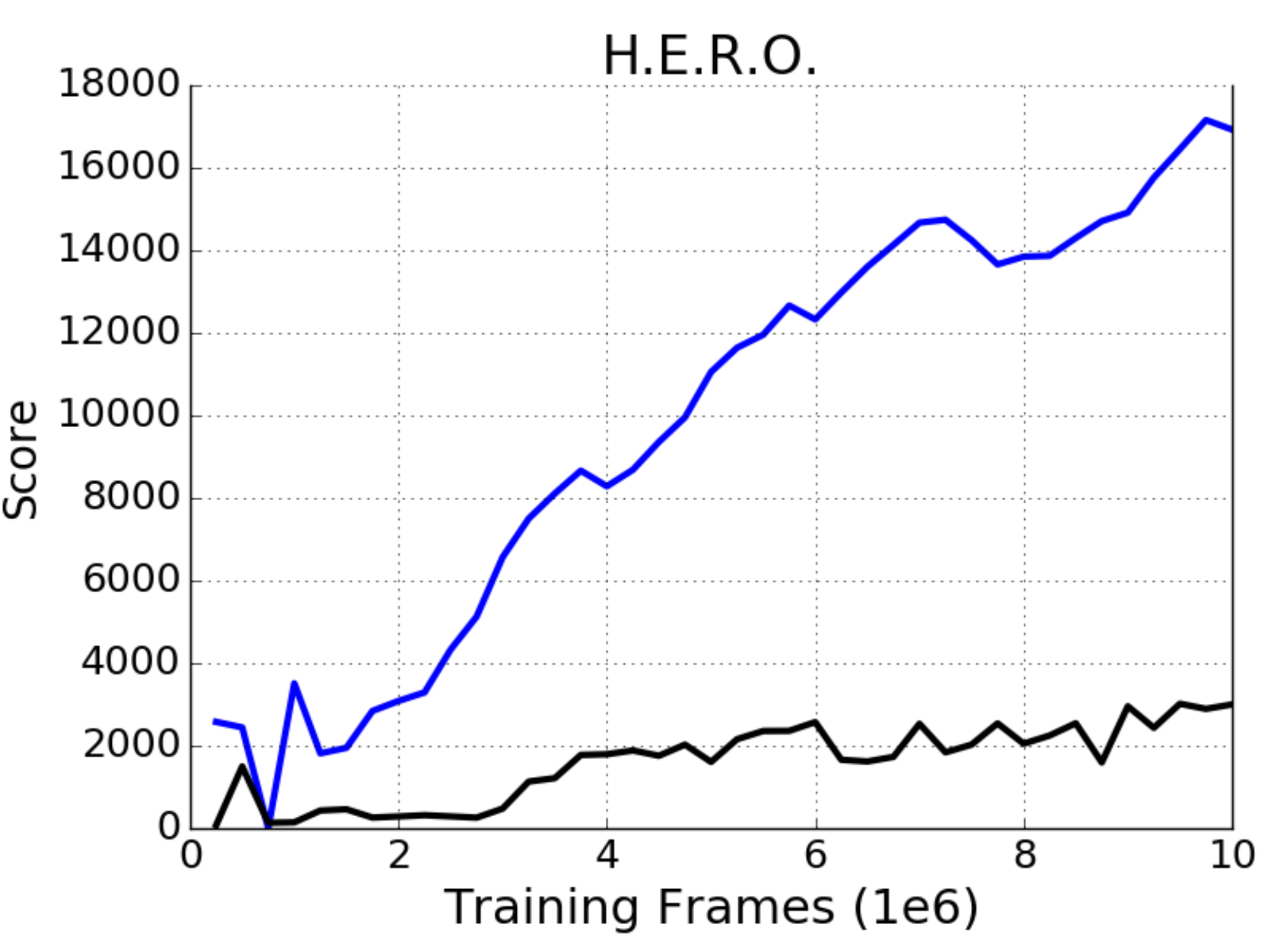}&
	 \includegraphics[width=0.3\textwidth]{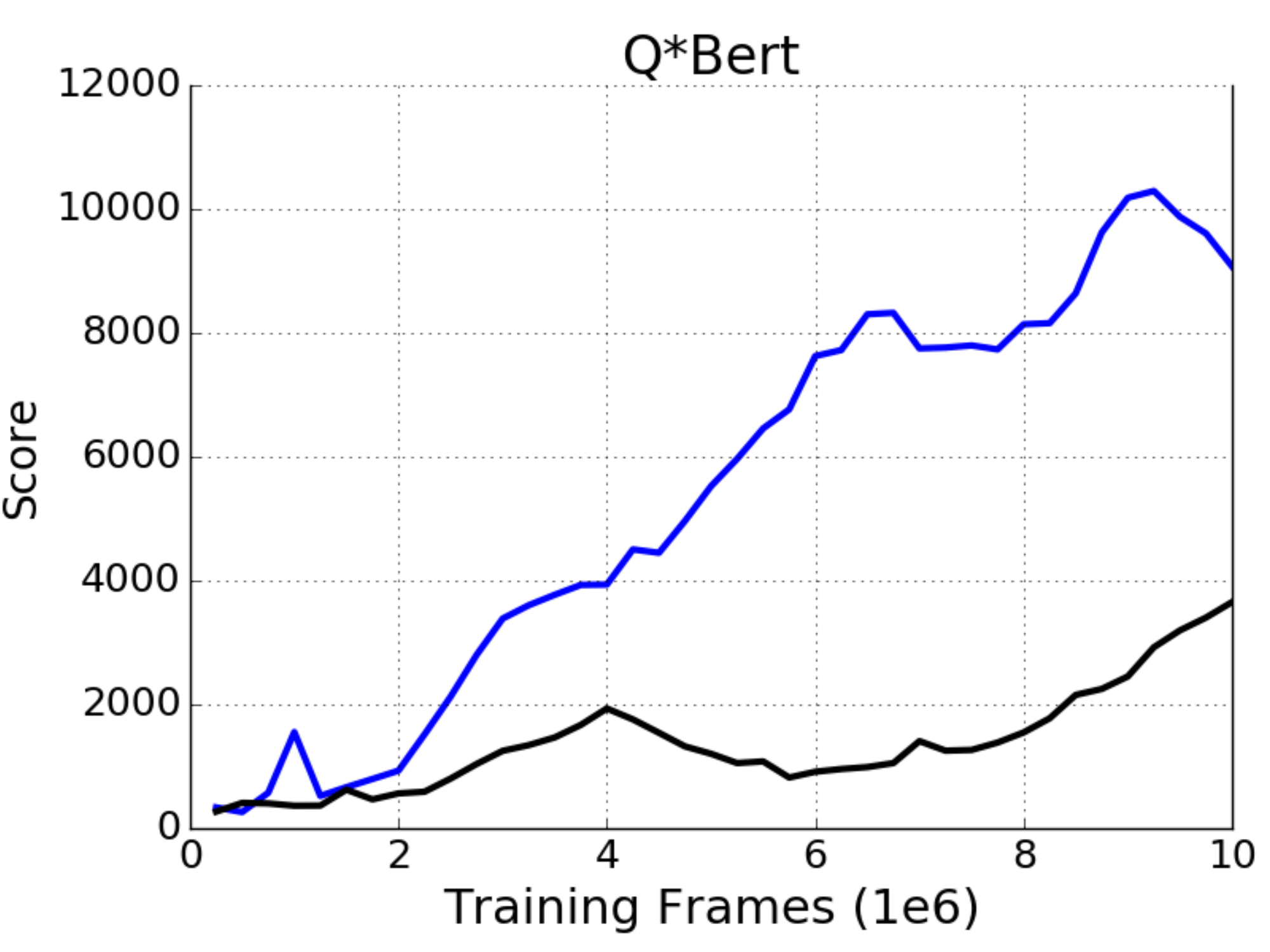}&
	 \includegraphics[width=0.3\textwidth]{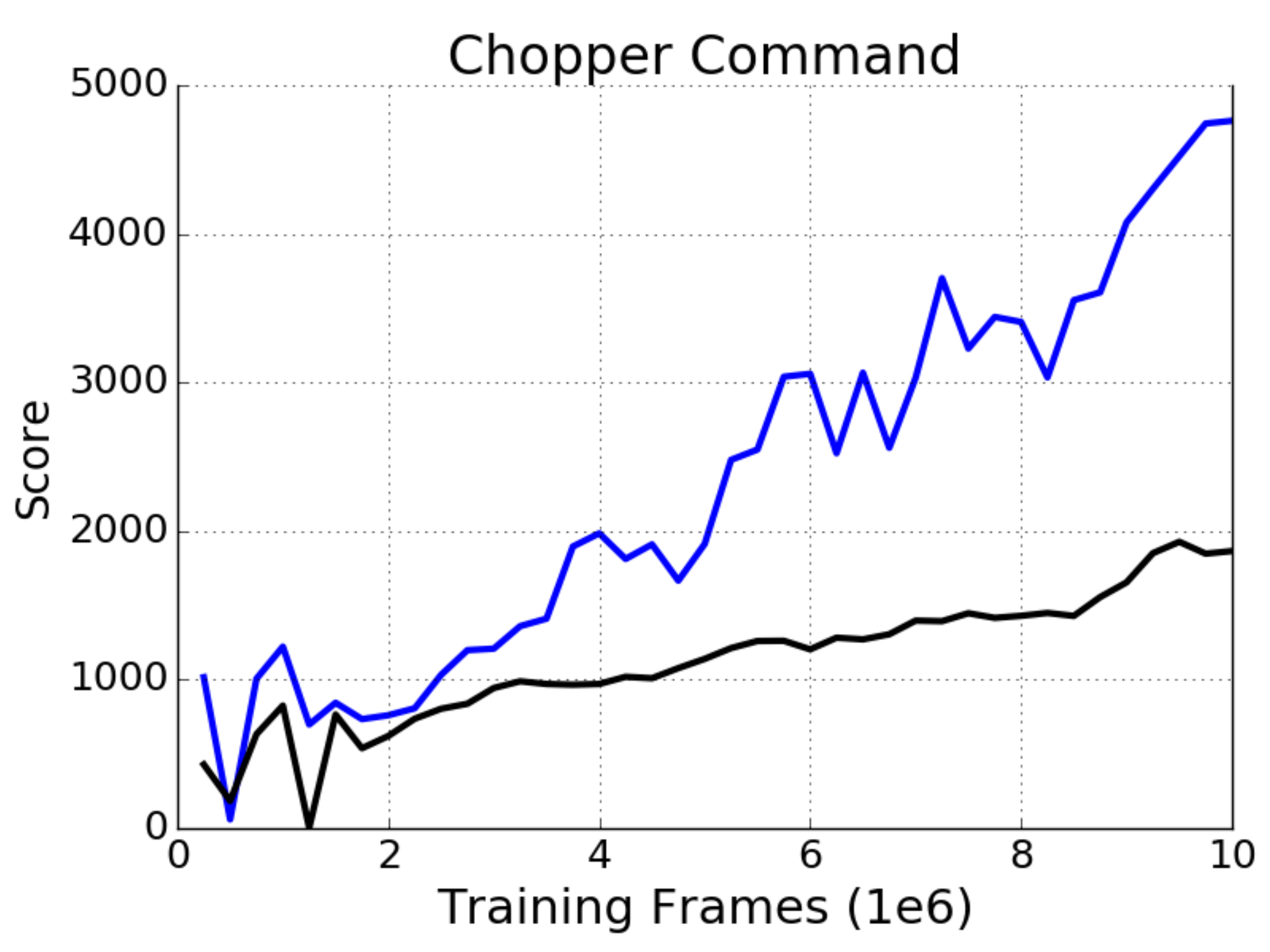}
    \end{tabular}
        \caption{Game scores for our algorithm (blue) and DQN (black) using 10M training frames. 30 no-op evaluation is used and moving average over 4 points is applied.}
    \label{fig:learning_curve1}
    \vspace{-0.3cm}
\end{figure}

In \figref{fig:learning_curve1} we illustrate the evolution of the score for our algorithm and the DQN approach for the 6 games `Frostbite,' `Atlantis,' `Zaxxon,' `H.E.R.O,' `Q*Bert,' and `Chopper Command.' We  observe our method to achieve significantly higher scores very early on. Importantly our technique  increases the gap between our approach and the  DQN performance  even during later stages of the training. 
We refer the reader to the supplementary material for additional results and raw scores.

\section{Conclusion}
In this paper we proposed a novel program for deep Q-learning which propagates promising rewards to achieve significantly faster convergence than the classical DQN. Our method significantly outperforms competing approaches even when trained on a small fraction of the data on the Atari 2600 domain.
In the future, we plan to investigate the impact of  penalty functions, advanced constrained optimization techniques and explore potential synergy with other techniques.

\pagebreak
{\small
\begingroup
\setlength{\bibsep}{2.5pt}
\bibliography{iclr2017_conference}
\bibliographystyle{iclr2017_conference}
\endgroup
}

\pagebreak
\appendix
\section{Supplementary Material}
\renewcommand{\thefigure}{S\arabic{figure}}
\renewcommand{\thetable}{S\arabic{table}}   
\setcounter{figure}{0}
\setcounter{table}{0}

\begin{figure}[H]
    \centering
\includegraphics[width=\linewidth]{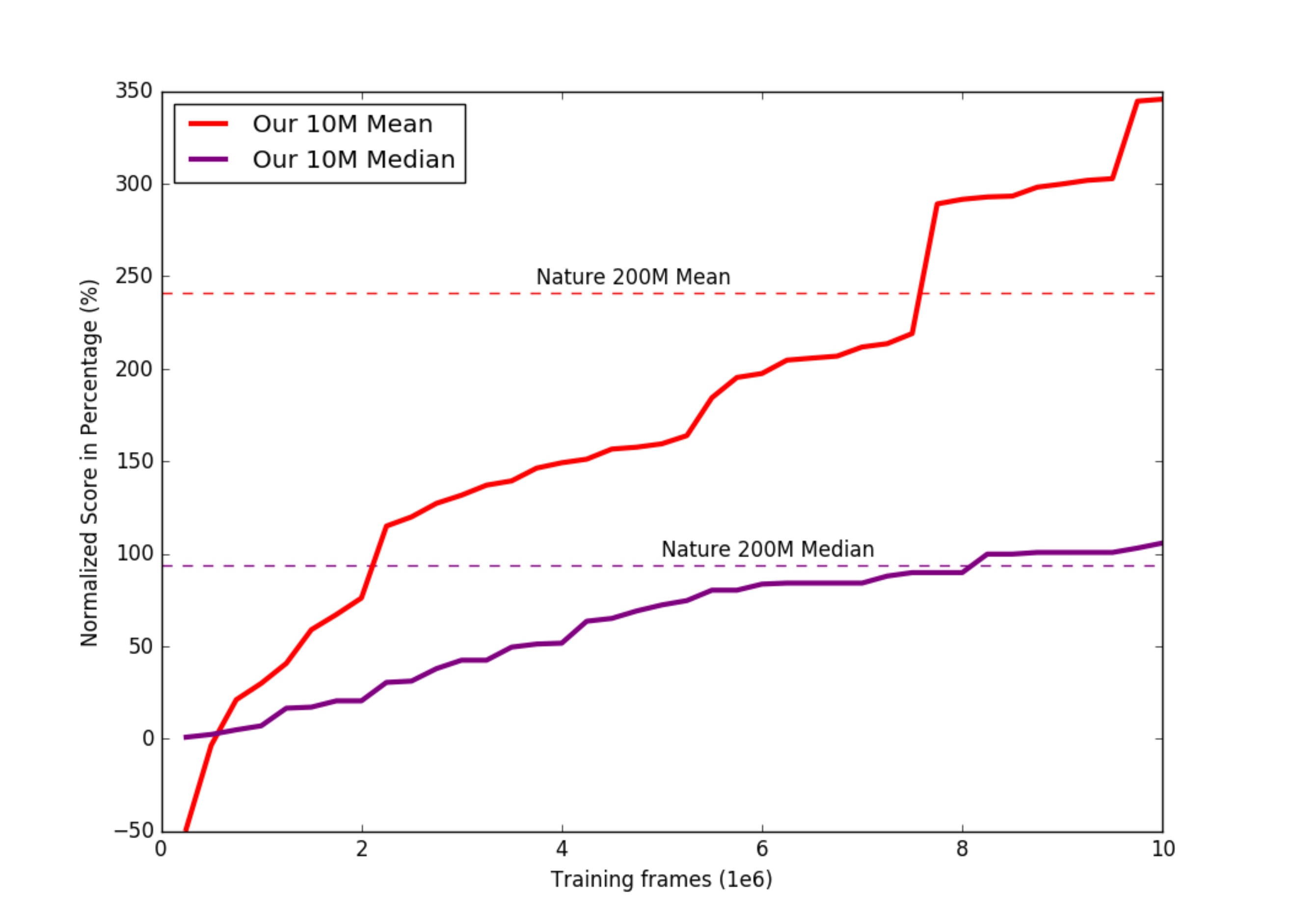}
    \caption{Convergence of mean and median of normalized percentages on 49 games.}    \label{fig:Normalized_mean_median}

\end{figure}

We present our quantitative results  in \tabref{tab:raw_score_table} and \tabref{tab:normalized_results_table}. We also illustrate the normalized score provided in \equref{eq:evaluation3} over the number of episodes in \figref{fig:Normalized_mean_median}.

\begin{table}[H]
    \centering
    \begin{tabular}{l c c c c}
        \textbf{Game} & \textbf{Random} & \textbf{Human} & \textbf{DQN 200M} & \textbf{Ours 10M} \\
        Alien               & 227.80    & 6875      & 3069      & 1864      \\
        Amidar              & 5.8       & 1676      & 739.5     & 565.67    \\
        Assault             & 222.4     & 1496      & 3359      & 5142.37   \\
        Asterix             & 210       & 8503      & 6012      & 5408.33   \\
        Asteroids           & 719.1     & 13157     & 1629      & 1481.67   \\
        Atlantis            & 12850     & 29028     & 85641     & 316766.67 \\
        Bank Heist          & 14.2      & 734.4     & 429.7     & 596       \\
        Battle Zone         & 2360      & 37800     & 26300     & 30800     \\
        Beam Rider          & 363.9     & 5775      & 6846      & 8069      \\
        Bowling             & 23.1      & 154.8     & 42.4      & 49.3      \\
        Boxing              & 0.1       & 4.3       & 71.8      & 81.17     \\
        Breakout            & 1.7       & 31.8      & 401.2     & 229.79    \\
        Centipede           & 2091      & 11963     & 8309      & 4470.06   \\
        Chopper Command     & 811       & 9882      & 6687      & 6360      \\
        Crazy Climber       & 10781     & 35411     & 114103    & 114146    \\
        Demon Attack        & 152.1     & 3401      & 9711      & 5738.67   \\
        Double Dunk         & -18.6     & -15.5     & -18.1     & -10.07    \\
        Enduro              & 0         & 309.6     & 301.8     & 672.83    \\
        Fishing Derby       & -91.7     & 5.5       & -0.8      & 5.27      \\
        Freeway             & 0         & 29.6      & 30.3      & 31.3      \\
        Frostbite           & 65.2      & 4335      & 328.3     & 3974.11   \\
        Gopher              & 257.6     & 2321      & 8520      & 4660      \\
        Gravitar            & 173       & 2672      & 306.7     & 346.67    \\
        H.E.R.O             & 1027      & 25763     & 19950     & 19975     \\
        Ice Hockey          & -11.2     & 0.9       & -1.6      & -3.43     \\
        Jamesbond           & 29        & 406.7     & 576.7     & 1088.33   \\
        Kangaroo            & 52        & 3035      & 6740      & 11716.67  \\
        Krull               & 1598      & 2395      & 3805      & 9461.1    \\
        Kung-Fu Master      & 258.5     & 22736     & 23270     & 27820     \\
        Montezuma's Revenge & 0         & 4376      & 0         & 23.33     \\
        Ms. Pacman          & 307.3     & 15693     & 2311      & 1805      \\
        Name This Game      & 2292      & 4076      & 7257      & 7314.67   \\
        Pong                & -20.7     & 9.3       & 18.9      & 19.4      \\
        Private Eye         & 24.9      & 69571     & 1788      & 342.37    \\
        Q*Bert              & 163.9     & 13455     & 10596     & 12355     \\
        River Raid          & 1339      & 13513     & 8316      & 8028.33   \\
        Road Runner         & 11.5      & 7845      & 18257     & 29346.67  \\
        Robotank            & 2.2       & 11.9      & 51.6      & 34.5      \\
        Seaquest            & 68.4      & 20182     & 5286      & 4070      \\
        Space Invaders      & 148       & 1652      & 1976      & 995       \\
        Star Gunner         & 664       & 10250     & 57997     & 16653.95  \\
        Tennis              & -23.8     & -8.9      & -2.5      & -1        \\
        Time Pilot          & 3568      & 5925      & 5947      & 5423.33   \\
        Tutankham           & 11.4      & 167.6     & 186.7     & 232       \\
        Up and Down         & 533.4     & 9082      & 8456      & 14406     \\
        Venture             & 0         & 1188      & 380       & 286.67    \\
        Video Pinball       & 16257     & 17298     & 42684     & 74873.2   \\
        Wizard of Wor       & 563.5     & 4757      & 3393      & 4716.67   \\
        Zaxxon              & 32.5      & 9173      & 4977      & 10598     \\
    \end{tabular}
    \caption{Raw Scores across 49 games, using 30 no-op start evaluation (5 minutes emulator time, 18000 frames, $\epsilon=$ 0.05). Results of DQN is taken from \citet{MnihNature2015} }
    \label{tab:raw_score_table}
\end{table}

\begin{table}[H]
    \centering
    \begin{tabular}{l c c}
        \textbf{Game} & \textbf{DQN 200M} & \textbf{Ours 10M} \\
        Alien               & 42.74\%   & 24.62\%   \\
        Amidar              & 43.93\%   & 33.52\%   \\
        Assault             & 246.27\%  & 386.31\%  \\
        Asterix             & 69.96\%   & 62.68\%   \\
        Asteroids           & 7.32\%    & 6.13\%    \\ 
        Atlantis            & 449.94\%  & 1878.60\% \\
        Bank Heist          & 57.69\%   & 80.78\%   \\
        Battle Zone         & 67.55\%   & 80.25\%   \\
        Beam Rider          & 119.79\%  & 142.39\%  \\
        Bowling             & 14.65\%   & 19.89\%   \\
        Boxing              & 1707.14\% & 1930.24\% \\
        Breakout            & 1327.24\% & 757.77\%  \\
        Centipede           & 62.99\%   & 24.10\%   \\
        Chopper Command     & 64.78\%   & 61.17\%   \\
        Crazy Climber       & 419.50\%  & 419.67\%  \\
        Demon Attack        & 294.22\%  & 171.95\%  \\
        Double Dunk         & 16.13\%   & 275.16\%  \\
        Enduro              & 97.48\%   & 217.32\%  \\
        Fishing Derby       & 93.52\%   & 99.76\%   \\
        Freeway             & 102.36\%  & 105.74\%  \\
        Frostbite           & 6.16\%    & 91.55\%   \\
        Gopher              & 400.43\%  & 213.36\%  \\
        Gravitar            & 5.35\%    & 6.95\%    \\
        H.E.R.O             & 76.50\%   & 76.60\%   \\
        Ice Hockey          & 79.34\%   & 64.22\%   \\
        Jamesbond           & 145.00\%  & 280.47\%  \\
        Kangaroo            & 224.20\%  & 391.04\%  \\
        Krull               & 276.91\%  & 986.59\%  \\
        Kung-Fu Master      & 102.38\%  & 122.62\%  \\
        Montezuma's Revenge & 0\%       & 0.53\%    \\
        Ms. Pacman          & 13.02\%   & 9.73\%    \\
        Name This Game      & 278.31\%  & 281.54\%  \\
        Pong                & 132\%     & 133.67\%  \\
        Private Eye         & 2.54\%    & 0.46\%    \\
        Q*Bert              & 78.49\%   & 91.73\%   \\
        River Raid          & 57.31\%   & 54.95\%   \\
        Road Runner         & 232.92\%  & 374.48\%  \\
        Robotank            & 509.28\%  & 332.99\%  \\
        Seaquest            & 25.94\%   & 19.90\%   \\
        Space Invaders      & 121.54\%  & 56.31\%   \\
        Star Gunner         & 598.10\%  & 166.81\%  \\
        Tennis              & 142.95\%  & 153.02\%  \\
        Time Pilot          & 100.93\%  & 78.72\%   \\
        Tutankham           & 112.23\%  & 141.23\%  \\
        Up and Down         & 92.68\%   & 162.38\%  \\
        Venture             & 31.99\%   & 24.13\%   \\
        Video Pinball       & 2538.62\% & 5630.76\% \\
        Wizard of Wor       & 67.47\%   & 99.04\%   \\
        Zaxxon              & 54.09\%   & 115.59\%  \\
    \end{tabular}
    \caption{Normalized results across 49 games, using the evaluation score given in \equref{eq:evaluation3}}
    \label{tab:normalized_results_table}
\end{table}

\end{document}